\documentclass[conference]{IEEEtran}
\IEEEoverridecommandlockouts

\usepackage{draftwatermark}
\SetWatermarkText{Preprint}
\SetWatermarkScale{0.12}
\SetWatermarkLightness{0.85}

\usepackage{cite}
\usepackage{amsmath,amssymb,amsfonts}
\usepackage{algorithm, algorithmic}
\usepackage{graphicx}
\usepackage{textcomp}
\usepackage{xcolor}
\def\BibTeX{{\rm B\kern-.05em{\sc i\kern-.025em b}\kern-.08em
    T\kern-.1667em\lower.7ex\hbox{E}\kern-.125emX}}
\begin{document}

\title{Local Performance vs. Out-of-Distribution Generalization: An Empirical Analysis of Personalized Federated Learning in Heterogeneous Data Environments}

\author{
\IEEEauthorblockN{Mortesa Hussaini, Jan Theiß, and Anthony Stein}
\IEEEauthorblockA{\textit{Computational Science Hub (CSH) \& Dept. Artificial Intelligence in Agricultural Engineering} \\
\textit{University of Hohenheim}\\
Stuttgart, Germany \\
\{mortesa.hussaini, jan.theiss, anthony.stein\}@uni-hohenheim.de\\ ORCID: \{0000-0002-3621-7776, 0009-0000-9112-5834, 0000-0002-1808-9758\}}
}

\maketitle

\begin{center}
\vspace{0.3cm}
\fbox{\parbox{0.9\columnwidth}{
\centering
\small
\textbf{This work is a preprint version and has been submitted in a slightly modified version to the IEEE for possible publication. \\
Copyright may be transferred without notice, after which this version may no longer be accessible.}
}}
\end{center}
\vspace{0.5cm}

\begin{abstract}
In the context of \textit{Federated Learning} with \textit{heterogeneous} data environments, local models tend to converge to their own local model optima during local training steps, deviating from the overall data distributions.
Aggregation of these local updates, e.g., with \textit{FedAvg}, often does not align with the global model optimum (\textit{``client drift"}), resulting in an update that is suboptimal for most clients.
\textit{Personalized Federated Learning} approaches address this challenge by exclusively focusing on the average \textit{local performances} of clients' models on their own data distribution.
\textit{Generalization} to out-of-distribution samples, which is a substantial benefit of FedAvg and represents a significant component of robustness, appears to be inadequately incorporated into the assessment and evaluation processes.
This study involves a thorough evaluation of Federated Learning approaches, encompassing both their local performance and their generalization capabilities.
Therefore, we examine different stages within a single communication round to enable a more nuanced understanding of the considered metrics.
Furthermore, we propose and incorporate a modified approach of FedAvg, designated as ``Federated Learning with Individualized Updates" \textit{(FLIU)}, extending the algorithm by a straightforward individualization step with an adaptive personalization factor.
We evaluate and compare the approaches empirically using MNIST and CIFAR-10 under various distributional conditions, including benchmark IID and pathological non-IID, as well as additional novel test environments with Dirichlet distribution specifically developed to stress the algorithms on complex data heterogeneity.
\end{abstract}


\section{Introduction}\label{sec:Intro}

Machine Learning (ML) relies on a large and diverse amount of data in order to adequately train reliable complex models, such as DNNs, and thus achieve a high level of predictive \textit{accuracy} and \textit{robustness} \cite{Lei_2019_HowTrainingData}.
\textit{Generalization} to out-of-distribution data is an aspect of robustness that we want to explore further here.
\textit{Federated Learning} (FL) is a distributed ML paradigm following the principle of bringing the model to the data and not vice versa.
Data-driven ML-models can be trained in a \textit{decentralized} way without the need for a central data-holding instance that needs direct access to all users' sensitive and private data.
Instead of potentially sensitive training data, only locally trained model parameters are exchanged.
FL can thereby obtain a global model with a high level of generalization by aggregating models of all participating clients.
Even though, high diversity in the training data theoretically leads to high generalization, FedAvg \cite{Mcmahan_2017_CommunicationEfficient} can lead to some challenges due to the often highly heterogeneous underlying data distributions \cite{Wen_2022_FLSurveyCA}, e.g. the \textit{client-drift} phenomena \cite{Karimireddy_2020_ClientDrift}.
The shared aggregated global model is not satisfactory for all clients since it doesn't represent client-specific local data properties well enough, which results in low \textit{local performance}.
Approaches, such as \cite{Deng_2020_APFL}, \cite{Zhang_2021_FedFOMO}
address this challenge by different techniques of \textit{Personalized Federated Learning} (PFL) \cite{Mansour_2020_PersonalizationFL}.
However, proper generalization should continue to be aspired, especially for clients with insufficient amount of data or new joining participants.
While some works briefly consider generalization effects \cite{sun_2024_generalization}, only few provide a unified metric that combines both aspects into a single evaluation framework.
Furthermore, a significant number of PFL approaches exhibit a lack of adaptability in the degree to which personalization is enforced across clients. 

This work addresses this gap by means of an analysis of both local performance and out-of-distribution generalization in FL under varying degrees of data heterogeneity.
In section \ref{sec:FLIU}, we define our optimization problem, derived from the one introduced in \cite{Mcmahan_2017_CommunicationEfficient} and propose \textit{Federated Learning with Individualized Updates} (FLIU), a simple yet effective extension of FedAvg incorporating an adaptive personalization step.
It is designed to achieve a balance between the sharing of global knowledge and sustaining personalized knowledge based on a clients individual data distribution.
For the sake of reproducibility, our experimental setup is delineated in detail (cf. Sec. \ref{sec:Experimental Setup}).
This includes own constructed non-IID data environments derived from the Dirichlet distribution \cite{Reguieg_dirichlet_FL_PFL_comp} and with label and quantity skew, thus enabling a more nuanced evaluation of local performance against out-of-distribution generalization.
Finally, we present our results and discuss them in detail (cf. Sec. \ref{sec:Res_and_Disc}) considering our introduced evaluation metrics and give an brief outlook about further research questions and possible solutions (cf. Sec. \ref{sec:Conclusion}).

\section{Related Work}\label{sec:Related Work}

Some FL Methods, e.g., FedProx \cite{Li_2020_FedProx}, introduce proximal terms to stabilize training in the presence of data heterogeneity.
Similarly, SCAFFOLD \cite{Karimireddy_2020_ClientDrift} uses variance reduction through control variates to address client drift.
While these methodologies intend to enhance global model convergence, they do not directly address performance on local distributions.
PFL aims to reconcile the tension between generalization and local performance.
A comprehensive taxonomy of PFL methods have emerged, encompassing meta-learning-based techniques such as Per-FedAvg \cite{Per-FedAvg_Fallah_2020}, model partitioning methods such as FedPer \cite{FedPer}, and regularization-based approaches such as Ditto \cite{Ditto_Li_2021}.
MiniPFL \cite{FAN_2024_MiniPFL} introduces a hierarchical framework, clustering clients into mini-federations based on the similarity of their model representations to reduce communication overhead while maintaining competitive accuracy under non-IID conditions.
Similarly, FedMPM \cite{Gan_2023_FedMPM} presents a mixed multi-stage personalization strategy that combines independently trained local models with historical global-aware models and effectively aligns local model updates and improves generalization across diverse data distributions.
\cite{Mansour_2020_PersonalizationFL} proposes model interpolation by blending of global and a local model using a learned or fixed coefficient.
However, this necessitates the training of the global model to identify the optimal interpolation between the client model and the global model.
This insight formalizes the intuition that local models tend to overfit small datasets, while global models may be misaligned with user distributions.
This is further substantiated by \cite{Hanzely_2021_FLMixture}, which proposes a similar optimization objective that explicitly encourages a mixture of local and global models through a quadratic penalty formulation, with formal convergence guarantees and reduced communication complexity.
FL+DE \cite{Peterson_2019_PrivateFLDomain} progresses this field by integrating the MoE framework, unifying differentially private global models with noise-free private models.
A gating function dynamically weights the influence of each, enabling smooth adaptation.
FedFOMO \cite{Zhang_2021_FedFOMO} has a slightly different technique, using only a smaller part of the federation for updating one client local model.
APFL \cite{Deng_2020_APFL} follows this principle by introducing a framework where each client maintains and updates a convex mixture of a global and a local model.
Notably, they derive generalization bounds dependent on the mixing parameter, which is then optimized adaptively based on empirical gradients. 
Their framework highlights the dynamic and client-specific nature of personalization, an idea we extend in our work through a more lightweight personalization step and alternative evaluation design.
Surveys \cite{Huang_2024_Survey_PFL_Challenges} and benchmarks, such as pFL-Bench \cite{pfl-bench_Chen_2022}, have systematically compared these approaches across a range of datasets and heterogeneity levels.

Extant literature reveals that no single method dominates across all evaluation metrics.
Trade-offs exist between local performance, generalization, communication cost and fairness. 
It is imperative to note that the preponderance of these assessments places considerable emphasis on the performance of local tests on clients' individual distributions.
However, this approach does not consider the exploration of generalization to unseen or skewed distributions.
FLIU is a PFL approach that modifies FedAvg to account for the heterogeneity across clients.
Furthermore, we present a novel perspective on the evaluation of PFL in standard benchmarks and recently developed heterogeneous data environments.

\section{FLIU}\label{sec:FLIU}

In this study, we examine the behavior of a modified version of the standard FedAvg aggregation method, namely FLIU, incorporating an additional personalization step on various \textit{personalization factors} and under different scenarios.
\begin{figure}[h]
    \centering
    \includegraphics[width=0.8\linewidth]{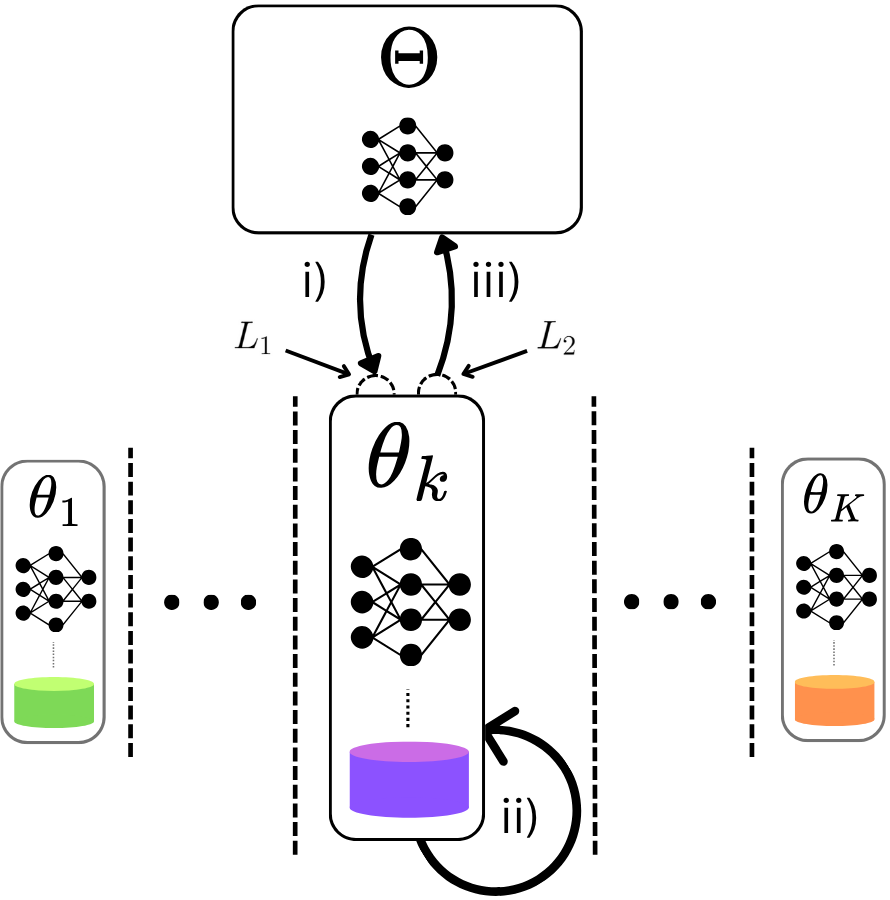}
    \caption{Schematic illustration of FLIU}
    \label{fig:platzhalter}
\end{figure}
For better understanding and comparability with other approaches, we delineate several stages, $G$, $L_1$ and $L_2$ (cf. Fig. \ref{fig:platzhalter}).
In stage $G$ the aggregation of the local model updates, through FedAvg, happens.
This updated global model is then been shared to the clients.
In stage $L_1$ the obtained global together with the trained local model from the previous round are used to give the client a personalized updated model.
In this instance, a comparison with FedAvg is pertinent, as this is the stage at which clients receive the new global model from the server.
In a real-world scenario where the model must be implemented immediately without the possibility of additional local optimization, the performance attained in this instance is of significance.
After local training through several local epochs, we reach stage $L_2$.
The distinction is that, in the interim, the client has utilized the individual updated model from stage $L_1$ with local samples, consequently yielding disparate test results.
Since, a considerable number of PFL methodologies consider solely stage $L_2$ for the purpose of evaluation, we consider this stage to compare FLIU with these PFL approaches.
To ensure the comparability of our model with all other models and to facilitate a more precise analysis of the model in this regard, both stages are considered here.

    \subsection{Notation \& Definition}

Let $C$ be the set of clients, $K = |C|$ the number of clients with $D_k$ the $k$-th client's data set, and $n_k = |D_k|$ the amount of samples from client $k \in C$ following an individual data distribution $P_{D_k}$.
Furthermore, let $D = \dot\bigcup_{k \in C} D_k$ be the overall data set with $n = |D| = \sum_{k \in C} |D_k|$ samples, each paired as $\left( x_i,y_i \right)$ with $i \in \{1,\ldots,n\}$.
For the sake of simplicity, we here assume horizontal FL \cite{IEEE_2021_GuideFML} with homogeneous models and a participation rate of $r=1.0$.
Going along with the notation as proposed in \cite{Mcmahan_2017_CommunicationEfficient}, we consider the optimization problem
\begin{equation}
\begin{split}
    \min_{\Theta \in \mathbb{R}^d} \sum_{k=1}^K \frac{n_k}{n} F_k \left( \Theta \right) \label{eq:FLProblem} \\
    \text{with} \, F_k \left( \Theta \right) = \frac{1}{n_k} \sum_{i=1}^{n_k} \ell \left( x_i,y_i,\Theta \right) \, ,
\end{split}
\end{equation}
where $\ell \left( \cdot \right)$ represents an arbitrary loss function and $\Theta \in \mathbb{R}^d$ the parameters of the global model, whose optimization is sought.
This optimization problem aims to find an optimal $\Theta^*$ such that the model performs best in the entire data set $D$.
Although this approach works fine in almost all IID and a few non-IID cases, it has been found to struggle to find an optimal global solution for $\Theta^*$, if the clients' local optimal parameters $\theta^*_k$ diverge too much from the globally optimal $\Theta^*$. This phenomenon is called client drift in the literature.
In addition, in some cases, $\Theta^*$ may be optimal in the overall data set $D$ or distribution $P_D$ but does not perform well on some client data sets $D_k$ or distribution $P_{D_k}$, respectively, worse than a locally trained model $\theta_k \in \mathbb{R}^d$.
For this reason, we propose a slight variation of the optimization problem \eqref{eq:FLProblem}, in which the focus is not exclusively on optimizing the global model $\Theta$ but rather on optimizing the individual local models $\theta_k \, , \forall \, k \in C$.
Nonetheless, the generalization capability of the global model $\Theta$, which is achieved by aggregating all client models $\theta_k$, should be further pursued, especially to offer robustness to clients whose database $n_k$ would be insufficient for exclusively local training.
Accordingly, our approach aims to optimize the following problem:
\begin{equation}
\begin{split}
    \min_{\theta_1,\ldots,\theta_K,\Theta \in \mathbb{R}^d} \sum_{k=1}^K \frac{n_k}{n} \left( F_k \left( \theta_k \right) + F_k \left( \Theta \right) \right) \label{eq:OurProblem} \\ \text{with} \quad F_k \left( \cdot \right) = \frac{1}{n_k} \sum_{i=1}^{n_k} \ell \left( x_i,y_i,\cdot \right) \, .
\end{split}
\end{equation}

    \subsection{Individualized Local Updates}

We introduce an approach with two steps:
First ($i)$) updating the global model $\Theta$ as usual for better generalization and thus robustness, and second to adapt it to the individual client conditions represented in their local data sets $D_k$.
Considering set $C$ of clients participating in the overall federation, we propose a straightforward individual update formula for any client $k \in C$, described by a linear combination
\begin{equation}
    \theta^{t+1}_k \leftarrow  \gamma_k * \theta^t_k + (1-\gamma_k) * \Theta^t \label{eq:aggregation} \, ,
\end{equation}
where $\Theta$ represents the global model, aggregated through the average of all client models, $\theta^t_k$ denotes the locally trained model ($L_2$) at communication round $t$ and $\gamma_k$ a weighting factor to obtain a personalized individual update for client $k$.
Even though, the selection of $\gamma_k$ depends on the use case, we have defined and implemented a function
\begin{equation}
    \gamma_k \left( n_k \right) = 
    \begin{cases}
        0.9 &, \text{if} \quad n_k > \frac{10n}{K} \\
        0.75 &, \text{if} \quad n_k > \frac{5n}{K} \\
        0.5 &, \text{if} \quad n_k > \frac{n}{K} \\
        0.25 &, \text{if} \quad n_k > \frac{n}{2K} \\
        0.1 &, \text{else}
    \end{cases}\label{eq:gamma}
\end{equation}
that adaptively adjusts to the relative amount of data for each client.
We suggest that a higher local training sample size induces higher robustness and thus less necessity for the respective client's model to obtain this through the global model. Algorithm \ref{alg:algorithm} summarizes FLIU in pseudo-code.

\begin{algorithm}
\caption{Let $C$ be the set of clients $k$ and $K = |C|$ the number of clients. $E$ denotes the local epochs. $\Theta$ the global model parameters and $\theta_k$ the local model parameters of client $k$. $\Theta^0$ is the initial model.}
\label{alg:algorithm}
\textbf{Server:}
\begin{algorithmic} 
\FOR{each client $k \in C$} 
\STATE initialize $\theta_k^0$ as $\Theta^0$
\ENDFOR
\vspace{0.5em}
\STATE \textbf{Communication:}
\FOR{each round $t= 1,2, \ldots$}
\FOR{each client $k \in C$}
\STATE ClientTraining
\ENDFOR
\STATE receive models $\theta_1^t, \ldots, \theta_K^t$ 
\vspace{0.3em}
\STATE $\Theta^t \leftarrow \sum_{kn\in C}  \frac{1}{K} \theta_k$
\vspace{0.3em}
\FOR{each client $k \in C$}
\STATE ClientUpdate($\Theta^t$)
\ENDFOR
\ENDFOR
\end{algorithmic}

\vspace{1em}
\textbf{Client:}
\begin{algorithmic}
\STATE \textbf{ClientTraining:}
\FOR{each local epoch $i \in \{1,\ldots,E\}$}
\STATE $\theta_k^t \leftarrow$ trainModel ($D_k,\ell(n_k,\theta_k^t),$AdamOptimizer)
\ENDFOR
\STATE send $\theta_k^t$ to server

\vspace{1em}
\STATE \textbf{ClientUpdate($\Theta^t$)}
\STATE $\theta^{t+1}_k \leftarrow  \gamma_k(n_k) * \theta^t_k + (1-\gamma_k(n_k)) * \Theta^t$
\end{algorithmic}
\end{algorithm}

\section{Experimental Setup}\label{sec:Experimental Setup}

This section outlines the structure of our experiments and provides models and experimental hyperparameter as well as a derivation of our own data environments and evaluation metrics, which we use in this work.

    \subsection{Data Environments}

Literature often distinguishes between different forms of distortion or skewness \cite{Li_2022_FLNonIID}
.
\textit{Label skew} considers the variation of the distribution of the labels $P_{D_k}(y_i)$ between clients.
Quantity skew describes the variation of the distribution in the size of the datasets $P_{D_k}(n_k = |D_k|)$ and thus, training samples between the clients.
For our experiments, we consider benchmark (cf. $1)$ \& $2)$) as well as own designed (cf. $3)$ - $5)$) data environments.

    \subsubsection{IID}\label{sec:iid}

A homogeneous data environment, where each client has approximately the same amount of training samples with all labels represented equally.

    \subsubsection{Pathological Non-IID}\label{sec:path_noniid}

One benchmark setting for label skew is the so called \textit{pathological} non-IID setting \cite{Mcmahan_2017_CommunicationEfficient}. 
Here, each client gets assigned two classes randomly among the 10 total classes of MNIST or CIFAR-10. 
To achieve this, we divide our data set $D$ into its ten classes and randomly assign just two classes to each client. This case depicts a label skew setting.
It simulates a highly heterogeneous data environment, since distributions $P_{D_k}$ can overlap completely (both classes are the same), partly (one class is the same) or not at all (no classes are same).
It is notably that we cannot guarantee any overlap between clients ($\exists k,l \in C : P_{D_k} \subset P_{D_l}$) for very small number of clients ($K \leq 10$).

    \subsubsection{Label-skew with Dirichlet Distribution (LS)}\label{sec:ls}
    
Here, we attain a non-IID setting with label-skew and constant $n_k, \, \forall k \in C$ through the Dirichlet distribution followed by an iterative proportional fitting using \textit{Sinkhorn-Knopps algorithm} \cite{sinkhorn1967concerning}.
\begin{figure}[h]
    \centering
    \includegraphics[width=\linewidth]{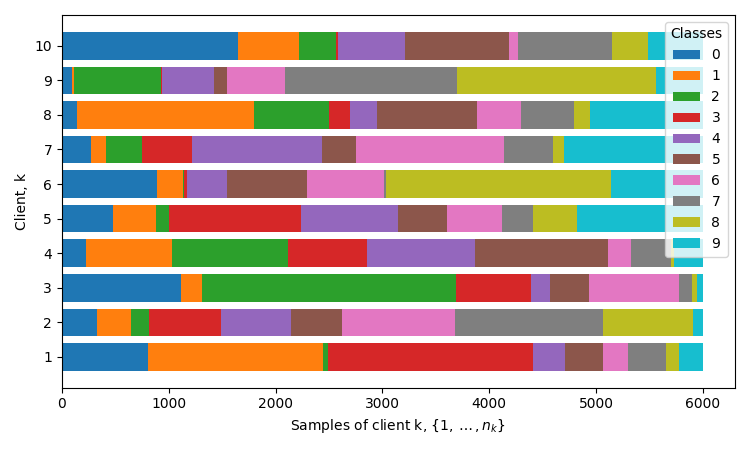}
    \caption{Example of a Dirichlet distributed ($\alpha = \mathbf{1}$) allocation skew of labels to $K=10$ clients from CIFAR-10.}
    \label{fig:dist_labelskew}
\end{figure}
Let $L$ be the number of classes of dataset $D$ and $P \in \mathbb{R}^{K \times L}$ a distribution matrix, where each row represents the distribution $p_k$ of classes for client $k$ with $p_k \sim {Dir} \left( \alpha_1,\ldots, \alpha_s \right)$ based on a Dirichlet distribution. The parameter $\alpha$ controls the level of skew in the distribution. A high $\alpha$ results in a more uniform distribution of labels across clients, while a low $\alpha$ creates a more skewed, imbalanced distribution.
After sampling, the matrix $P$ is normalized using the Sinkhorn-Knopp algorithm.
To retain its stochastic properties and to ensure all samples are distributed, we generate a matrix $P_S \approx P$, where the sum of each column (classes) equals one and the sum of each row (clients) equals $\frac{K}{L}$. This allows for a proportional distribution of data samples.
Using Sinkhorn-Knopp algorithm returns the unique matrix $P_S$ that minimizes the \textit{Kullback-Leibler divergence} under those mentioned row and column constraints \cite{Rüschendorf_1995}.
With this, we achieve an unbalanced data distribution with varying overlaps and which is closer to reality than the pathological non-IID setting, c.f. figure \ref{fig:dist_labelskew}.

    \subsubsection{Quantity-skew with Dirichlet Distribution (QS)}\label{sec:qs}

The quantity-skew is achieved by utilizing a draw from the Dirichlet distribution to be used as a quantity distribution of dataset sizes for each client: $\mathbf{P}_{D_k}(n_k = |D_k|) \sim {Dir} \left( \alpha_1,\ldots, \alpha_s \right) \, , \forall k \in C$.
\begin{figure}[h]
    \centering
    \includegraphics[width=\linewidth]{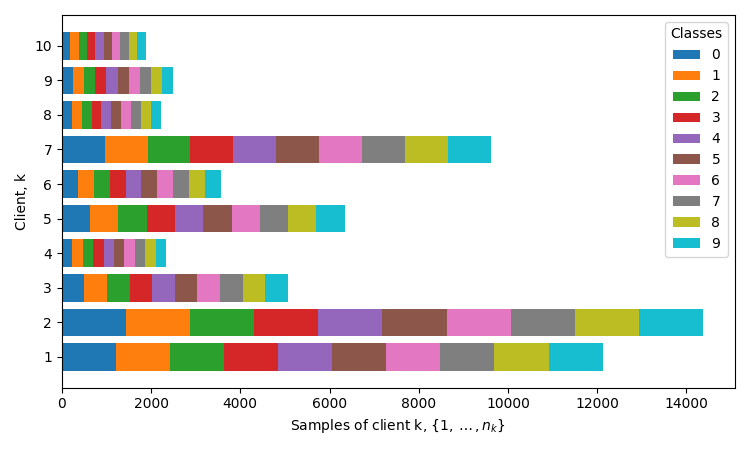}
    \caption{Example of a Dirichlet distributed ($\alpha = \mathbf{1}$) allocation skew of number of training samples to $K=10$ clients from CIFAR-10.}
    \label{fig:dist_qs}
\end{figure}
The class distribution within a clients dataset is balanced resulting in no label skew, but only a quantity skew across clients, c.f. figure \ref{fig:dist_qs}.
With that, we intend to identify the behavior and effects of unequal amounts of data (which also occur and are expected in reality) on the performance and robustness of individual client models.

    \subsubsection{Label- and Quantity-skew with Dirichlet Distribution (LSQS)}\label{sec:lsqs}

For our last setting, we combined label and quantity skew to evaluate the methods on a fully Dirichlet distributed environment.
\begin{figure}[h]
    \centering
    \includegraphics[width=\linewidth]{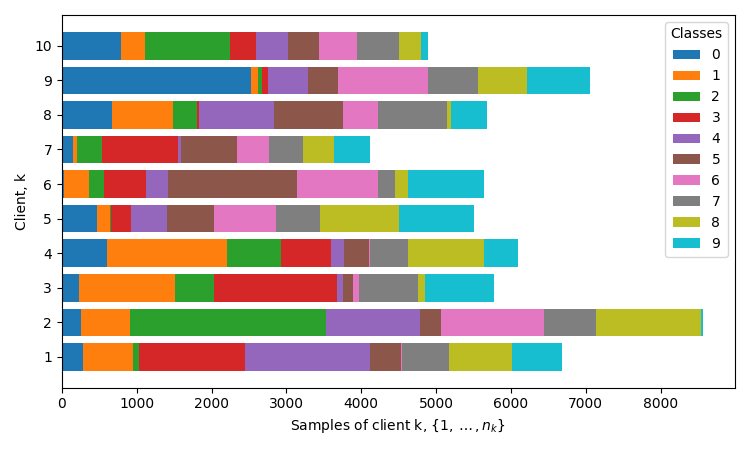}
    \caption{Example of a Dirichlet distributed ($\alpha = \mathbf{1}$) allocation skew of number of training samples and labels to $K=10$ clients from CIFAR-10.}
    \label{fig:dist_lsqs}
\end{figure}
This setting utilizes a Dirichlet distribution for both the overall training sample size but also the class label imbalance, c.f. figure \ref{fig:dist_lsqs}.
This scenario, with high heterogeneity with regard to amount of data and label distribution for each client, promises to come closest to reality.

    \subsection{Evaluation Metrics}

Our analysis encompasses two primary components.
Firstly, it assesses the performance of local models within their designated data environment.
Secondly, it evaluates the out-of-distribution generalization capability of these models across the overall data distribution.
It is imperative that we elucidate our comprehension of the term ``performance''.
In the context of classification tasks in PFL, average accuracy of local models within their respective data environment is the prevailing metric.
This metric can be regarded as a measure of \textit{local performance} $\mathbf{Acc}\left(L\right)$.
Nevertheless, this metric is inadequate for our research purposes because it doesn't assess the performance of these models under other data distributions.
Consequently, no conclusions can be drawn about the generalization capabilities.
To this end, the accuracy of the models is assessed when evaluated on an aggregate data set, constructed from the individual client test sets.
This assessment is declared as the \textit{generalization performance} $\mathbf{Acc}\left(G\right)$.
In order to take into account both metrics, it is necessary to consider $\mathbf{Acc} = \mathbf{Acc}\left(L\right) + \mathbf{Acc}\left(G\right)$.
To perform an exhaustive evaluation, it is necessary to assess stages $L_1$ and $L_2$.

We introduce $\rho_\epsilon = |\{k \in C: \mathbf{Acc}\left(L\right)_k > \epsilon\}|$ as a metric to count the number of clients whose local performance surpass a certain threshold $\epsilon$ to achieve additional insight about the local performance $\mathbf{Acc}\left(L\right)$.
It provides insights that extend beyond the average local performance across all clients.
In this context, a higher $\rho$ with higher $\epsilon$ indicates greater adaptability of the updates to the respective client's local conditions and their individual performances.

    \subsection{ML-Models and Experimental Hyperparameter}
    
To retain comparability, for MNIST, we use the simple CNN introduced in \cite{Mcmahan_2017_CommunicationEfficient}, which consists of five hidden layers: Two Convolutional Layers, each followed by one Max-Pooling Layer and at last a Fully-Connected Layer. 
Regarding CIFAR-10, the CNN-model is adjusted to the given input and an additional Convolutional Layer is added before the Fully-Connected Layer.
For our experiments, we consider $K \in \{10, 100\}$ clients and use a batch size of $B=50$, $E=5$ number of local epochs and $T=100$ communication rounds. 
For FedAvg and FLIU, the learning rate is set with $\eta=0.01$ for MNIST and $\eta=0.001$ for CIFAR-10 with a learning rate decay of 0.99.
Regarding our update formula \eqref{eq:aggregation} we analyze different personalization factors $\gamma_k \in \{0.1, 0.25, 0.5, 0.75, 0.9\}$ as well as the adaptive personalization function \eqref{eq:gamma}. 

APFL and FedFomo were selected for comparison due to their design combining models and thereby serving as a suitable benchmark for comparison.
To reproduce the training process for APFL, we use the setting presented in \cite{Deng_2020_APFL} with a learning rate $\eta=0.1$ for both MNIST and CIFAR-10
.
For FedFomo, we reproduced their experiments \cite{Zhang_2021_FedFOMO} with a learning rate $\eta=0.01$ for MNIST and $\eta=0.1$ for CIFAR-10 without a learning rate decay.
Further we set the number of downloaded models per client to 5.
We used PFLlib \cite{Zhang_2025_PFLlib} to reimplement APFL and FedFomo with the models described above.
Comparable to \cite{Zhang_2021_FedFOMO}, we reproduced our experiments $N=3$ times for statistical relevance.

\section{Results \& Discussion}\label{sec:Res_and_Disc}

In the initial phase of the study, we conduct an analysis of FLIU under various personalization factors $\gamma$.
This analysis was undertaken to ascertain the efficacy of the proposed approach in diverse (heterogeneous) data environments.
To this end, a comparative analysis was performed with FedAvg and centrally local training (CLT).
Subsequently, an additional comparison with APFL and FedFomo is presented.
This comparison encompasses not only local performance and out-of-distribution generalization, but also the number of clients undercutting or exceeding a certain performance threshold $\rho$.
The latter provides further insight into the efficacy of the methods, as it reveals the number of clients who benefit from them and the number of those who do not.
\begin{table*}[!t]
\caption{Average performances of $C=100$ clients evaluated on MNIST with different data distribution. Results averaged from $N=3$ repetitions, each with $T=100$ rounds.}
\label{tab:100_MNIST}
\centering
\begin{tabular}{|l|ccccccc|}
\hline
\textbf{MNIST} & $L_1$ $\mathbf{Acc}\left(L\right)$ & $L_1$ $\mathbf{Acc}\left(G\right)$ & $L_1$ $\mathbf{Acc}$ & $L_2$ $\mathbf{Acc}\left(L\right)$ & $L_2$ $\mathbf{Acc}\left(G\right)$ & $L_2$ $\mathbf{Acc}$ & G $\mathbf{Acc}\left(G\right)$ \\
\hline
\textbf{IID} &  &  &  &  &  &  & \\
CLT & 92.8 ± 1.12 & 92.8 ± 1.15 & 185.6 ± 2.26 & 92.8 ± 1.12 & 92.8 ± 1.15 & 185.6 ± 2.26 & - \\
FedAvg & 98.73 ± 0.07 & \textbf{98.73 ± 0.07} & \textbf{197.46 ± 0.14} & 98.05 ± 0.23 & 97.99 ± 0.23 & 196.04 ± 0.44 & \textbf{98.73 ± 0.07} \\
FLIU & 98.66 ± 0.09 & 98.61 ± 0.08 & 197.27 ± 0.17 & 98.05 ± 0.12 & 97.96 ± 0.11 & 196.01 ± 0.22 & 98.69 ± 0.04 \\
FLIU10 & 98.62 ± 0.11 & 98.64 ± 0.11 & 197.26 ± 0.23 & 97.92 ± 0.4 & 97.97 ± 0.3 & 195.89 ± 0.69 & 98.67 ± 0.1 \\
FLIU25 & \textbf{98.74 ± 0.12} & 98.71 ± 0.08 & 197.45 ± 0.2 & \textbf{98.21 ± 0.23} & \textbf{98.17 ± 0.23} & \textbf{196.39 ± 0.46} & \textbf{98.73 ± 0.1} \\
FLIU50 & 98.48 ± 0.03 & 98.48 ± 0.06 & 196.96 ± 0.07 & 98.06 ± 0.24 & 98.07 ± 0.23 & 196.14 ± 0.48 & 98.57 ± 0.07 \\
FLIU75 & 97.67 ± 0.26 & 97.64 ± 0.13 & 195.31 ± 0.38 & 97.3 ± 0.38 & 97.23 ± 0.26 & 194.53 ± 0.64 & 97.95 ± 0.07 \\
FLIU90 & 95.54 ± 0.23 & 95.41 ± 0.27 & 190.95 ± 0.5 & 95.28 ± 0.23 & 95.08 ± 0.29 & 190.36 ± 0.52 & 96.1 ± 0.3 \\
\hline
\textbf{PATH} &  &  &  &  &  &  & \\
CLT & 99.11 ± 0.31 & 19.59 ± 0.05 & 118.69 ± 0.36 & 99.11 ± 0.31 & 19.59 ± 0.05 & 118.69 ± 0.36 & - \\
FedAvg & 85.72 ± 1.42 & 85.84 ± 1.48 & 171.57 ± 2.9 & 99.44 ± 0.12 & 29.95 ± 7.05 & 129.39 ± 7.17 & 85.84 ± 1.48 \\
FLIU & 99.46 ± 0.14 & 81.84 ± 0.85 & 181.3 ± 0.99 & \textbf{99.69 ± 0.04} & 36.13 ± 1.78 & 135.82 ± 1.82 & 95.83 ± 1.13 \\
FLIU10 & 94.95 ± 1.42 & \textbf{89.36 ± 1.93} & 184.31 ± 3.32 & 99.49 ± 0.01 & 41.12 ± 0.52 & 140.61 ± 0.51 & 90.74 ± 2.29 \\
FLIU25 & 98.78 ± 0.67 & 88.1 ± 4.05 & \textbf{186.88 ± 4.68} & 99.66 ± 0.05 & \textbf{43.47 ± 7.06} & \textbf{143.14 ± 7.1} & 92.09 ± 5.03 \\
FLIU50 & \textbf{99.58 ± 0.09} & 64.39 ± 3.06 & 163.97 ± 2.98 & 99.58 ± 0.06 & 26.54 ± 0.63 & 126.12 ± 0.62 & \textbf{96.92 ± 0.2} \\
FLIU75 & 99.46 ± 0.14 & 30.73 ± 4.28 & 130.19 ± 4.28 & 99.32 ± 0.11 & 21.61 ± 1.05 & 120.93 ± 0.97 & 94.87 ± 0.69 \\
FLIU90 & 98.91 ± 0.08 & 19.54 ± 0.03 & 118.45 ± 0.11 & 99.03 ± 0.08 & 19.56 ± 0.02 & 118.59 ± 0.11 & 33.4 ± 6.88 \\
\hline
\textbf{LS} &  &  &  &  &  &  & \\
CLT & 90.42 ± 0.9 & 81.29 ± 1.35 & 171.72 ± 2.25 & 90.42 ± 0.9 & 81.29 ± 1.35 & 171.72 ± 2.25 & - \\
FedAvg & 98.35 ± 0.08 & \textbf{98.35 ± 0.08} & 196.7 ± 0.17 & 97.84 ± 0.27 & 97.02 ± 0.47 & 194.86 ± 0.74 & 98.35 ± 0.08 \\
FLIU & \textbf{98.48 ± 0.07} & 98.33 ± 0.16 & \textbf{196.81 ± 0.23} & 97.96 ± 0.34 & 97.18 ± 0.63 & 195.14 ± 0.97 & \textbf{98.43 ± 0.11} \\
FLIU10 & 98.25 ± 0.57 & 98.22 ± 0.58 & 196.48 ± 1.15 & 97.74 ± 0.83 & 96.86 ± 1.26 & 194.6 ± 2.09 & 98.24 ± 0.59 \\
FLIU25 & 98.44 ± 0.13 & \textbf{98.35 ± 0.16} & 196.79 ± 0.28 & \textbf{98.16 ± 0.07} & \textbf{97.46 ± 0.17} & \textbf{195.62 ± 0.23} & 98.4 ± 0.17 \\
FLIU50 & 98.32 ± 0.09 & 98.12 ± 0.1 & 196.43 ± 0.19 & 98.01 ± 0.12 & 97.26 ± 0.18 & 195.27 ± 0.28 & 98.31 ± 0.1 \\
FLIU75 & 97.59 ± 0.07 & 96.97 ± 0.26 & 194.55 ± 0.32 & 97.35 ± 0.14 & 96.17 ± 0.48 & 193.51 ± 0.62 & 97.61 ± 0.11 \\
FLIU90 & 93.82 ± 1.03 & 89.16 ± 2.59 & 182.98 ± 3.6 & 93.79 ± 0.91 & 88.12 ± 2.52 & 181.91 ± 3.41 & 88.77 ± 6.38 \\
\hline
\textbf{QS} &  &  &  &  &  &  & \\
CLT & 87.29 ± 0.57 & 86.99 ± 0.59 & 174.28 ± 1.12 & 87.29 ± 0.57 & 86.99 ± 0.59 & 174.28 ± 1.12 & - \\
FedAvg & \textbf{98.68 ± 0.18} & \textbf{98.8 ± 0.17} & \textbf{197.49 ± 0.35} & \textbf{98.38 ± 0.31} & 98.35 ± 0.3 & \textbf{196.73 ± 0.6} & \textbf{98.8 ± 0.17} \\
FLIU & \textbf{98.68 ± 0.09} & 98.71 ± 0.1 & 197.4 ± 0.19 & 98.34 ± 0.16 & 98.36 ± 0.19 & 196.7 ± 0.35 & 98.78 ± 0.08 \\
FLIU10 & 98.65 ± 0.29 & 98.79 ± 0.1 & 197.43 ± 0.38 & 98.18 ± 0.38 & \textbf{98.38 ± 0.16} & 196.55 ± 0.53 & \textbf{98.8 ± 0.11} \\
FLIU25 & 98.66 ± 0.06 & 98.78 ± 0.04 & 197.43 ± 0.03 & 98.22 ± 0.12 & 98.34 ± 0.08 & 196.56 ± 0.2 & \textbf{98.8 ± 0.05} \\
FLIU50 & 98.47 ± 0.15 & 98.59 ± 0.08 & 197.07 ± 0.21 & 98.04 ± 0.28 & 98.17 ± 0.24 & 196.22 ± 0.5 & 98.71 ± 0.05 \\
FLIU75 & 97.73 ± 0.07 & 97.85 ± 0.09 & 195.58 ± 0.14 & 97.42 ± 0.1 & 97.51 ± 0.12 & 194.94 ± 0.22 & 98.15 ± 0.08 \\
FLIU90 & 93.19 ± 0.71 & 93.53 ± 0.47 & 186.73 ± 1.18 & 93.07 ± 0.65 & 93.12 ± 0.47 & 186.19 ± 1.12 & 93.56 ± 1.02 \\
\hline
\textbf{LSQS} &  &  &  &  &  &  & \\
CLT & 89.25 ± 1.67 & 78.95 ± 1.92 & 168.2 ± 3.59 & 89.25 ± 1.67 & 78.95 ± 1.92 & 168.2 ± 3.59 & - \\
FedAvg & \textbf{98.81 ± 0.17} & \textbf{98.81 ± 0.16} & \textbf{197.62 ± 0.32} & 98.18 ± 0.24 & 97.38 ± 0.53 & 195.57 ± 0.77 & \textbf{98.81 ± 0.16} \\
FLIU & 98.61 ± 0.17 & 98.47 ± 0.3 & 197.09 ± 0.46 & 97.99 ± 0.17 & 97.42 ± 0.47 & 195.41 ± 0.64 & 98.57 ± 0.28 \\
FLIU10 & 98.53 ± 0.42 & 98.49 ± 0.44 & 197.02 ± 0.86 & 97.91 ± 0.33 & 96.95 ± 0.78 & 194.86 ± 1.11 & 98.51 ± 0.45 \\
FLIU25 & 98.68 ± 0.06 & 98.64 ± 0.06 & 197.32 ± 0.08 & \textbf{98.23 ± 0.15} & \textbf{97.66 ± 0.19} & \textbf{195.89 ± 0.26} & 98.69 ± 0.05 \\
FLIU50 & 98.55 ± 0.08 & 98.36 ± 0.01 & 196.91 ± 0.08 & 98.19 ± 0.02 & 97.47 ± 0.22 & 195.66 ± 0.22 & 98.55 ± 0.02 \\
FLIU75 & 97.82 ± 0.18 & 97.11 ± 0.33 & 194.93 ± 0.51 & 97.45 ± 0.35 & 96.14 ± 0.48 & 193.59 ± 0.83 & 97.95 ± 0.14 \\
FLIU90 & 94.47 ± 0.39 & 90.62 ± 0.73 & 185.09 ± 1.11 & 94.15 ± 0.34 & 89.35 ± 0.76 & 183.5 ± 1.1 & 93.97 ± 0.7 \\
\hline
\end{tabular}
\end{table*}
As previously outlined, the $L_1$ stage is of significance as clients are provided with the updated model here, for the purpose of conducting further tasks.
It is reasonable to hypothesize that models will demonstrate effective performance at this stage, since clients generally seek models with high local performance without the necessity to first conduct local training in order to adapt it to their own environment and conditions.
Regrettably, this aspect is frequently overlooked in literature.

Tables \ref{tab:100_MNIST} and \ref{tab:100_CIFAR} present the mean results of the performance metrics of $K=100$ clients across all stages ($L_1$, $L_2$ and $G$) evaluated on MNIST and CIFAR-10 with varied distribution configurations.
FedAvg demonstrates robust performance overall.
However, it reveals challenges in dealing with exceedingly high label skewness in PATH, where FLIU seems to outperform it.
Generally, FLIU shows consistently high performance across various environmental contexts, often ranking among the top performers in certain scenarios.
Our results suggest that FedAvg demonstrates a distinct enhancement in performance when operating in less extreme label skew environments, as depicted by the Dirichlet distribution and that PATH is not a fair setup for FedAvg.
FLIU has demonstrated consistent resilience across diverse environmental contexts.
Nonetheless, the utilization of a high fixed individualization factor $\gamma$ (here $\in {0.75, 0.9}$) appears to be suboptimal, as the update factor is excessively dependent on the samples trained on the local models' optima, as evidenced by the suboptimal performance of CLT in all of the considered environments.

\begin{figure}[h]
    \centering
    {\includegraphics[width=\linewidth]{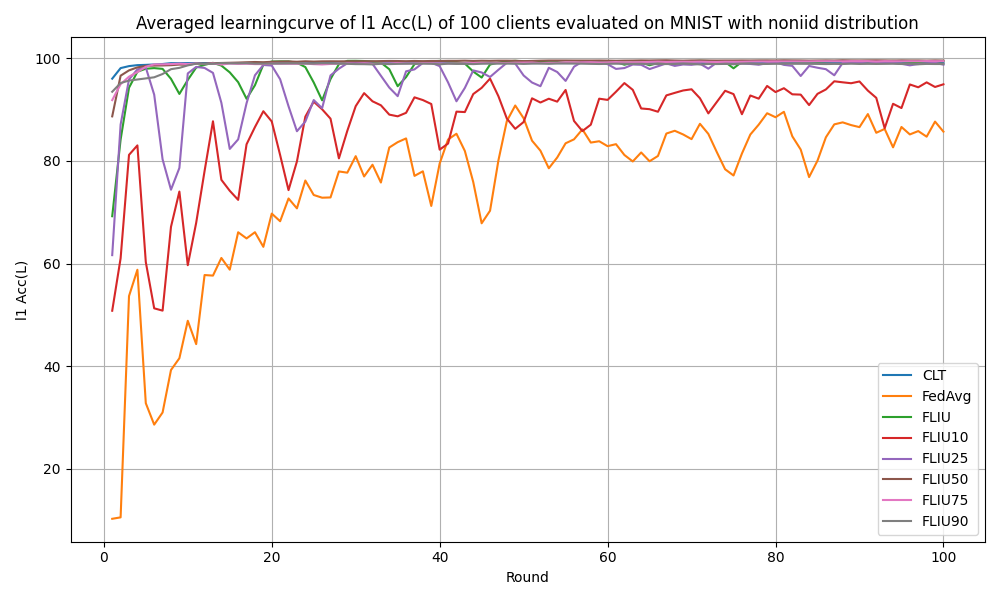}}
    \caption{Comparison of the trajectories of $\mathbf{Acc}\left(L\right)$ at stage $L_1$ of CLT, FedAvg and FLIU with adaptive $\gamma_k$ and several fixed $\gamma$s on MNIST with PATH for T=100 rounds.}\label{fig:L1_Local_Performance}
\end{figure}
\begin{figure}[h]
    \centering
    {\includegraphics[width=\linewidth]{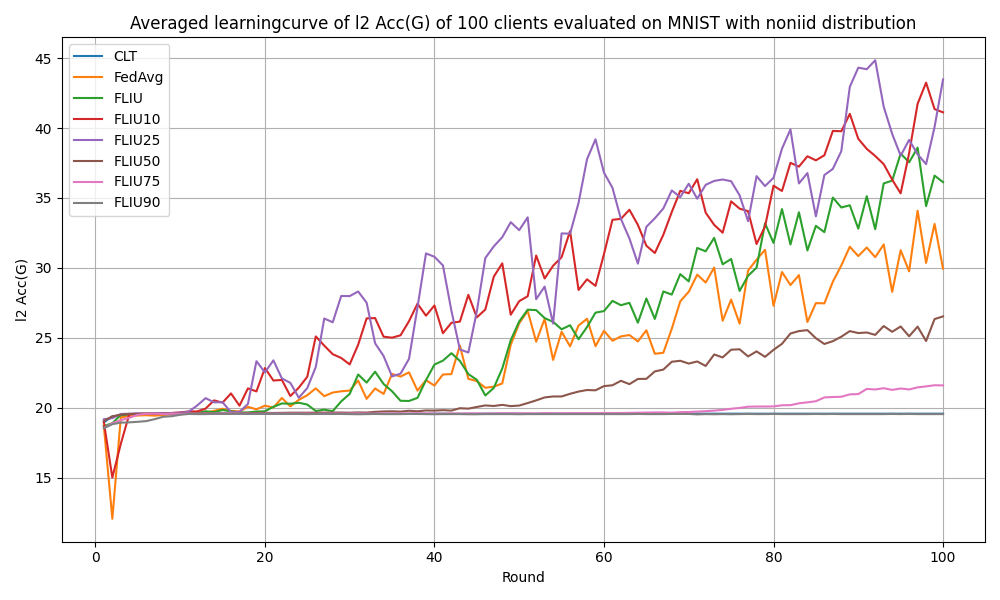}}
    \caption{Comparison of the trajectories of $\mathbf{Acc}\left(G\right)$ at stage $L_2$ of CLT, FedAvg and FLIU with adaptive $\gamma_k$ and several fixed $\gamma$s on MNIST with PATH for T=100 rounds.}\label{fig:L2_Generalization}
\end{figure}

Tables \ref{tab:100_MNIST} and \ref{tab:100_CIFAR} also provide insight into the performance discrepancy between the stages $L_1$ and $L_2$.
It appears that, on a high level (which is attained expeditiously on MNIST due to its simplicity), further local training epochs diminish performance in nearly all environments (possibly through overfitting or catastrophic forgetting \cite{catastrophic}).
Nevertheless, given the increased complexity associated with PATH, the performance of the models do not attain such elevated levels in this context.
As we know, the local model benefits from higher out-of-distribution generalization when the global model is adopted into the update.
However, as a trade-off, it loses significant local performance directly after updating.
Nevertheless, Figure  \ref{fig:L1_Local_Performance} shows that, despite losing in every round, $\mathbf{Acc}\left(L\right)$ increases over several rounds in stage $L_1$.
An observation of the same kind, but in the opposite direction, can be made in stage $L_2$.
Training solely on local samples during several local epochs greatly improves increase the model's local performance, but in exchange, decrease its ability to generalize to out-of-distribution data.
Here, too, we can observe in figure \ref{fig:L2_Generalization} that $\mathbf{Acc}\left(L\right)$ increases over several rounds in stage $L_2$.
Overall, we see that FLIU effectively captures the losses of local performance and out-of-distribution generalization at the different stages, balancing them against a higher value to ultimately achieve a higher performance level in both metrics.
FLIU demonstrates robust performance across a range of environments, even in scenarios where FedAvg experiences challenges.
Furthermore, FLIU consistently attains high scores throughout all phases of the evaluation process, particularly also in phase $G$, with the performance of the global model, which is important considering new clients entering the federation and initially receiving this model.

\begin{table*}[!t]
\caption{Average performances of $K=100$ clients evaluated on CIFAR-10 with different data distribution. Results averaged from $N=3$ repetitions, each with $T=100$ rounds.}
\label{tab:100_CIFAR}
\centering
\begin{tabular}{|l|ccccccc|}
\hline
\textbf{CIFAR10} & $L_1$ $\mathbf{Acc}\left(L\right)$ & $L_1$ $\mathbf{Acc}\left(G\right)$ & $L_1$ $\mathbf{Acc}$ & $L_2$ $\mathbf{Acc}\left(L\right)$ & $L_2$ $\mathbf{Acc}\left(G\right)$ & $L_2$ $\mathbf{Acc}$ & G $\mathbf{Acc}\left(G\right)$ \\
\hline
\textbf{IID} &  &  &  &  &  &  & \\
CLT & 36.15 ± 0.15 & 36.19 ± 0.21 & 72.34 ± 0.35 & 36.15 ± 0.15 & 36.19 ± 0.21 & 72.34 ± 0.35 & 23.61 ± 5.04 \\
FedAvg & \textbf{64.83 ± 0.45} & \textbf{64.83 ± 0.45} & \textbf{129.65 ± 0.89} & 58.83 ± 0.44 & 59.07 ± 0.45 & 117.9 ± 0.88 & \textbf{64.83 ± 0.45} \\
APFL & - & - & - & \textbf{60.5 ± 0.78} & \textbf{60.64 ± 0.96} & \textbf{121.15 ± 1.74} & 64.04 ± 0.62 \\
FedFomo & - & - & - & 37.05 ± 3.29 & 36.97 ± 3.44 & 74.02 ± 6.72 & - \\
FLIU & 59.33 ± 1.07 & 59.32 ± 1.01 & 118.65 ± 2.09 & 52.69 ± 1.61 & 52.84 ± 1.26 & 105.53 ± 2.86 & 59.86 ± 1.02 \\
FLIU10 & 63.52 ± 1.28 & 63.46 ± 1.43 & 126.98 ± 2.7 & 57.23 ± 1.7 & 57.18 ± 1.32 & 114.41 ± 3.0 & 63.58 ± 1.47 \\
FLIU25 & 60.6 ± 0.88 & 60.78 ± 0.77 & 121.38 ± 1.64 & 53.98 ± 0.94 & 54.3 ± 0.94 & 108.28 ± 1.88 & 61.32 ± 0.78 \\
FLIU50 & 53.28 ± 0.71 & 53.27 ± 0.77 & 106.55 ± 1.46 & 48.09 ± 0.56 & 48.05 ± 0.77 & 96.14 ± 1.32 & 55.36 ± 0.59 \\
FLIU75 & 49.28 ± 0.48 & 48.98 ± 0.45 & 98.27 ± 0.89 & 47.02 ± 0.38 & 46.53 ± 0.45 & 93.55 ± 0.68 & 53.05 ± 0.4 \\
FLIU90 & 43.94 ± 1.88 & 44.07 ± 1.87 & 88.01 ± 3.74 & 42.94 ± 1.87 & 43.11 ± 1.85 & 86.05 ± 3.72 & 47.53 ± 2.33 \\
\hline
\textbf{PATH} &  &  &  &  &  &  & \\
CLT & 85.71 ± 0.03 & 16.98 ± 0.01 & 102.69 ± 0.03 & 85.71 ± 0.03 & 16.98 ± 0.01 & 102.69 ± 0.03 & 17.09 ± 2.48 \\
FedAvg & 31.05 ± 3.51 & 31.05 ± 3.51 & 62.09 ± 7.02 & \textbf{87.96 ± 0.53} & \textbf{17.48 ± 0.07} & \textbf{105.44 ± 0.61} & 31.05 ± 3.51 \\
APFL & - & - & - & 70.48 ± 1.19 & 13.99 ± 0.27 & 84.47 ± 1.45 & 18.7 ± 3.55 \\
FedFomo & - & - & - & 52.68 ± 0.6 & 11.59 ± 0.02 & 64.27 ± 0.61 & - \\
FLIU & 71.68 ± 3.08 & 31.91 ± 1.5 & 103.59 ± 1.71 & 87.09 ± 0.31 & 17.31 ± 0.02 & 104.4 ± 0.29 & 36.72 ± 0.77 \\
FLIU10 & 48.25 ± 6.48 & \textbf{33.9 ± 4.51} & 82.15 ± 10.72 & 87.73 ± 0.42 & 17.44 ± 0.07 & 105.17 ± 0.49 & 34.9 ± 5.01 \\
FLIU25 & 67.54 ± 2.55 & 28.67 ± 2.23 & 96.21 ± 2.55 & 87.28 ± 0.33 & 17.31 ± 0.03 & 104.59 ± 0.35 & 32.72 ± 4.15 \\
FLIU50 & 85.69 ± 0.52 & 23.13 ± 0.69 & \textbf{108.82 ± 1.19} & 86.04 ± 0.42 & 17.12 ± 0.11 & 103.16 ± 0.53 & 44.87 ± 1.08 \\
FLIU75 & \textbf{86.32 ± 0.81} & 17.7 ± 0.12 & 104.02 ± 0.92 & 86.15 ± 0.82 & 17.08 ± 0.13 & 103.23 ± 0.96 & \textbf{45.61 ± 1.79} \\
FLIU90 & 85.35 ± 0.31 & 16.94 ± 0.05 & 102.29 ± 0.36 & 85.38 ± 0.42 & 16.94 ± 0.06 & 102.31 ± 0.47 & 36.1 ± 0.31 \\
\hline
\textbf{LS} &  &  &  &  &  &  & \\
CLT & 48.39 ± 0.3 & 31.28 ± 0.1 & 79.67 ± 0.3 & 48.39 ± 0.3 & 31.28 ± 0.1 & 79.67 ± 0.3 & 27.67 ± 3.02 \\
APFL & - & - & - & 13.51 ± 1.53 & 11.12 ± 0.88 & 24.63 ± 2.42 & 10.71 ± 1.25 \\
FedFomo & - & - & - & 13.15 ± 1.67 & 10.5 ± 0.74 & 23.65 ± 2.38 & - \\
FedAvg & 62.31 ± 0.26 & \textbf{62.31 ± 0.26} & 124.62 ± 0.52 & \textbf{64.87 ± 0.49} & \textbf{49.63 ± 0.26} & \textbf{114.51 ± 0.73} & \textbf{62.31 ± 0.26} \\
FLIU & 59.86 ± 0.21 & 53.13 ± 0.16 & 112.99 ± 0.35 & 57.98 ± 0.3 & 43.78 ± 0.05 & 101.76 ± 0.32 & 56.18 ± 0.14 \\
FLIU10 & \textbf{63.55 ± 0.53} & 61.59 ± 0.68 & \textbf{125.14 ± 1.21} & 64.18 ± 0.93 & 48.96 ± 0.81 & 113.14 ± 1.74 & 61.8 ± 0.64 \\
FLIU25 & 62.53 ± 1.09 & 57.92 ± 0.97 & 120.46 ± 2.05 & 60.87 ± 0.96 & 46.21 ± 1.0 & 107.08 ± 1.96 & 58.95 ± 0.77 \\
FLIU50 & 59.56 ± 1.0 & 51.49 ± 1.19 & 111.04 ± 2.19 & 57.22 ± 0.98 & 43.18 ± 0.97 & 100.39 ± 1.95 & 55.6 ± 1.11 \\
FLIU75 & 56.96 ± 0.21 & 44.93 ± 0.73 & 101.89 ± 0.93 & 55.8 ± 0.6 & 41.14 ± 0.71 & 96.94 ± 1.31 & 52.05 ± 0.75 \\
FLIU90 & 52.82 ± 0.74 & 38.42 ± 0.82 & 91.23 ± 1.55 & 52.5 ± 0.76 & 36.85 ± 0.86 & 89.34 ± 1.62 & 46.41 ± 2.1 \\
\hline
\textbf{QS} &  &  &  &  &  &  & \\
CLT & 32.97 ± 1.12 & 32.57 ± 0.64 & 65.54 ± 1.6 & 32.97 ± 1.12 & 32.57 ± 0.64 & 65.54 ± 1.6 & 27.96 ± 4.74 \\
FedAvg & \textbf{67.19 ± 0.62} & \textbf{67.13 ± 0.79} & \textbf{134.32 ± 1.4} & \textbf{60.82 ± 0.63} & \textbf{60.64 ± 0.65} & \textbf{121.46 ± 1.28} & \textbf{67.13 ± 0.79} \\
APFL & - & - & - & 21.37 ± 4.83 & 21.21 ± 4.88 & 42.58 ± 9.71 & 21.26 ± 4.79 \\
FedFomo & - & - & - & 9.24 ± 0.65 & 9.03 ± 0.84 & 18.27 ± 1.48 & - \\
FLIU & 58.58 ± 1.74 & 58.26 ± 1.44 & 116.84 ± 3.17 & 52.1 ± 1.67 & 52.31 ± 1.56 & 104.4 ± 3.23 & 59.2 ± 1.46 \\
FLIU10 & 65.95 ± 0.8 & 65.69 ± 0.24 & 131.64 ± 1.03 & 59.73 ± 0.26 & 59.1 ± 0.37 & 118.83 ± 0.62 & 65.75 ± 0.2 \\
FLIU25 & 62.88 ± 0.71 & 62.65 ± 0.54 & 125.53 ± 1.23 & 56.18 ± 0.54 & 56.24 ± 0.59 & 112.42 ± 1.13 & 63.19 ± 0.55 \\
FLIU50 & 55.8 ± 1.09 & 55.67 ± 1.15 & 111.47 ± 2.24 & 51.2 ± 1.37 & 50.86 ± 1.06 & 102.05 ± 2.38 & 57.65 ± 1.28 \\
FLIU75 & 51.44 ± 1.28 & 51.63 ± 0.48 & 103.07 ± 1.75 & 49.1 ± 0.91 & 49.34 ± 0.43 & 98.44 ± 1.34 & 55.03 ± 0.5 \\
FLIU90 & 44.69 ± 0.86 & 44.51 ± 1.14 & 89.2 ± 1.98 & 43.67 ± 0.85 & 43.29 ± 1.11 & 86.96 ± 1.94 & 49.65 ± 1.65 \\
\hline
\textbf{LSQS} &  &  &  &  &  &  & \\
CLT & 47.26 ± 0.91 & 30.95 ± 0.57 & 78.2 ± 1.48 & 47.26 ± 0.91 & 30.95 ± 0.57 & 78.2 ± 1.48 & 25.36 ± 5.32 \\
FedAvg & 62.8 ± 0.34 & \textbf{62.8 ± 0.32} & \textbf{125.6 ± 0.65} & \textbf{65.14 ± 0.88} & \textbf{50.43 ± 0.41} & \textbf{115.58 ± 1.29} & \textbf{62.8 ± 0.32} \\
APFL  & - & - & - & 19.6 ± 1.84 & 13.57 ± 0.44 & 33.16 ± 2.28 & 13.97 ± 1.2 \\
FedFomo & - & - & - & 17.07 ± 0.19 & 12.52 ± 0.14 & 29.59 ± 0.33 & - \\
FLIU & 60.31 ± 2.03 & 54.11 ± 1.91 & 114.43 ± 3.94 & 58.47 ± 2.19 & 44.77 ± 1.8 & 103.25 ± 4.0 & 56.43 ± 1.91 \\
FLIU10 & \textbf{63.4 ± 1.18} & 61.76 ± 1.14 & 125.16 ± 2.31 & 63.77 ± 1.12 & 49.27 ± 0.94 & 113.04 ± 2.05 & 61.99 ± 1.25 \\
FLIU25 & 62.48 ± 0.58 & 58.16 ± 0.3 & 120.65 ± 0.72 & 60.8 ± 0.21 & 46.87 ± 0.4 & 107.68 ± 0.53 & 59.25 ± 0.5 \\
FLIU50 & 59.58 ± 0.93 & 51.86 ± 0.74 & 111.44 ± 1.64 & 57.08 ± 0.58 & 43.77 ± 0.76 & 100.85 ± 1.33 & 55.81 ± 0.88 \\
FLIU75 & 55.8 ± 1.34 & 44.59 ± 0.68 & 100.4 ± 2.02 & 54.65 ± 1.3 & 40.84 ± 0.63 & 95.49 ± 1.93 & 51.75 ± 0.93 \\
FLIU90 & 50.54 ± 0.67 & 36.47 ± 0.73 & 87.01 ± 1.39 & 50.21 ± 0.7 & 34.95 ± 0.69 & 85.17 ± 1.39 & 44.9 ± 1.05 \\
\hline
\end{tabular}
\end{table*}

In order to establish a basis for comparison with other PFL approaches, it is necessary to focus on after local finetuning at stage $L_2$, as this is the stage that is chosen in the majority of other works for evaluation.
\begin{table*}[!t]
\caption{Average performances of $C \in \{10, 100\}$ clients evaluated on MNIST and CIFAR-10 with respectively LS, QS and LSQS. Results averaged from $N=3$ repetitions, each with $T=100$ rounds.}
\label{tab:eval2}
\centering
\begin{tabular}{|l|cc|cc|cc|}
\hline
 & \multicolumn{2}{|c|}{LS} & \multicolumn{2}{|c|}{QS} & \multicolumn{2}{|c|}{LSQS} \\
\hline
\textbf{MNIST} & $L_2$ $\mathbf{Acc}\left(L\right)$ & $L_2$ $\mathbf{Acc}\left(G\right)$ & $L_2$ $\mathbf{Acc}\left(L\right)$ & $L_2$ $\mathbf{Acc}\left(G\right)$ & $L_2$ $\mathbf{Acc}\left(L\right)$ & $L_2$ $\mathbf{Acc}\left(G\right)$ \\
\hline
\textbf{C=10} &  &  &  &  &  &  \\
CLT & 95.34 ± 3.73 & 91.39 ± 4.62 & 90.93 ± 4.9 & 91.08 ± 5.0 & 91.01 ± 6.64 & 85.16 ± 8.97 \\
FedAvg & 98.66 ± 0.17 & 98.48 ± 0.22 & 98.83 ± 0.19 & 98.87 ± 0.18 & 98.47 ± 0.15 & 98.01 ± 0.34 \\
APFL & \textbf{99.28 ± 0.01} & \textbf{99.08 ± 0.0} & \textbf{99.14 ± 0.08} & \textbf{99.22 ± 0.02} & \textbf{99.32 ± 0.02} & \textbf{99.12 ± 0.01} \\
FedFomo & 93.8 ± 0.34 & 90.64 ± 0.35 & 79.07 ± 0.05 & 79.11 ± 0.03 & 94.51 ± 0.08 & 92.93 ± 0.15 \\
FLIU & 98.86 ± 0.07 & 98.77 ± 0.04 & 99.04 ± 0.07 & 99.04 ± 0.03 & 98.85 ± 0.14 & 98.73 ± 0.18 \\
\hline
\textbf{C=100} &  &  &  &  &  &  \\
CLT & 90.42 ± 0.9 & 81.29 ± 1.35 & 87.29 ± 0.57 & 86.99 ± 0.59 & 89.25 ± 1.67 & 78.95 ± 1.92 \\
FedAvg & 97.84 ± 0.27 & 97.02 ± 0.47 & \textbf{98.38 ± 0.31} & 98.35 ± 0.3 & \textbf{98.18} ± 0.24 & 97.38 ± 0.53 \\
APFL & 96.87 ± 0.11 & 94.23 ± 0.07 & 98.37 ± 0.01 & 98.22 ± 0.02 & 97.07 ± 0.04 & 95.21 ± 0.13 \\
FedFomo & 94.28 ± 0.37 & 92.27 ± 0.48 & 20.41 ± 0.85 & 20.4 ± 0.67 & 94.5 ± 0.28 & 92.04 ± 0.19 \\
FLIU & \textbf{97.96 ± 0.34} & \textbf{97.18 ± 0.63} & 98.34 ± 0.16 & \textbf{98.36 ± 0.19} & 97.99 ± 0.17 & \textbf{97.42 ± 0.47} \\
\hline
\textbf{CIFAR-10} &  &  &  &  &  &  \\
\hline
\textbf{C=10} &  &  &  &  &  &  \\
CLT & 63.04 ± 0.13 & 47.1 ± 0.94 & 48.13 ± 1.24 & 47.35 ± 0.55 & 64.87 ± 0.63 & 44.44 ± 0.39 \\
FedAvg & 71.87 ± 1.1 & 61.35 ± 0.74 & \textbf{71.01 ± 0.41} & \textbf{69.99 ± 0.39} & 73.91 ± 0.48 & 59.36 ± 0.52 \\
APFL & \textbf{72.83 ± 0.2} & \textbf{64.55 ± 0.08} & 69.76 ± 0.32 & 69.08 ± 0.27 & \textbf{74.64 ± 0.15} & \textbf{63.19 ± 0.02} \\
FedFomo & 22.95 ± 2.32 & 10.0 ± 0.0 & 10.0 ± 0.0 & 10.0 ± 0.0 & 24.23 ± 0.0 & 10.0 ± 0.0 \\
FLIU & 71.01 ± 1.18 & 60.33 ± 1.19 & 66.52 ± 1.25 & 65.89 ± 1.17 & 71.36 ± 0.51 & 56.87 ± 0.15 \\
\hline
\textbf{C=100} &  &  \\
CLT & 48.39 ± 0.3 & 31.28 ± 0.1 & 32.97 ± 1.12 & 32.57 ± 0.64 & 47.26 ± 0.91 & 30.95 ± 0.57 \\
FedAvg & \textbf{64.87 ± 0.49} & \textbf{49.63 ± 0.26} & \textbf{60.82 ± 0.63} & \textbf{60.64 ± 0.65} & \textbf{65.14 ± 0.88} & \textbf{50.43 ± 0.41} \\
APFL & 13.51 ± 1.53 & 11.12 ± 0.88 & 21.37 ± 4.83 & 21.21 ± 4.88 & 19.6 ± 1.84 & 13.57 ± 0.44 \\
FedFomo & 13.15 ± 1.67 & 10.5 ± 0.74 & 9.24 ± 0.65 & 9.03 ± 0.84 & 17.07 ± 0.19 & 12.52 ± 0.14 \\
FLIU & 57.98 ± 0.3 & 43.78 ± 0.05 & 52.1 ± 1.67 & 52.31 ± 1.56 & 58.47 ± 2.19 & 44.77 ± 1.8 \\
\hline
\end{tabular}
\end{table*}
In the context of a limited sample size of $K=10$ clients, APFL demonstrates superior performance across almost all metrics in both the MNIST and CIFAR-10 datasets, especially in the three custom-built scenarios (LS, QS, and LSQS) (cf. Tab. \ref{tab:eval2}).
Nevertheless, the discrepancy in comparison to FLIU is frequently negligible.
The case is even reversed when considering $K=100$ instead of $K=10$ clients.
It appears that APFL is encountering difficulties in maintaining its performance levels in this context, particularly in scenarios involving the more intricate CIFAR-10 dataset, where a substantial decline in performance is observed.
In the course of our experimental inquiries, FedFomo has demonstrated a deficiency in its performances even on the MNIST dataset.
It is imperative to reiterate that no hyperparameter optimization was conducted, whether for FLIU or the other models.
The hyperparameter for APFL and FedFomo were selected based on the parameters outlined in the respective papers.
Given the elevated level of intricacy inherent in the design and operational characteristics of these models, there is a possibility that they necessitate optimization in specific scenarios.
Nevertheless, empirical evidence suggests that FedAvg and FLIU demonstrate a superior capacity to manage the increased complexity inherent in CIFAR-10, without the necessity of adjusting any hyperparameter.
\begin{figure}[h]
    \centering
    {\includegraphics[width=\linewidth]{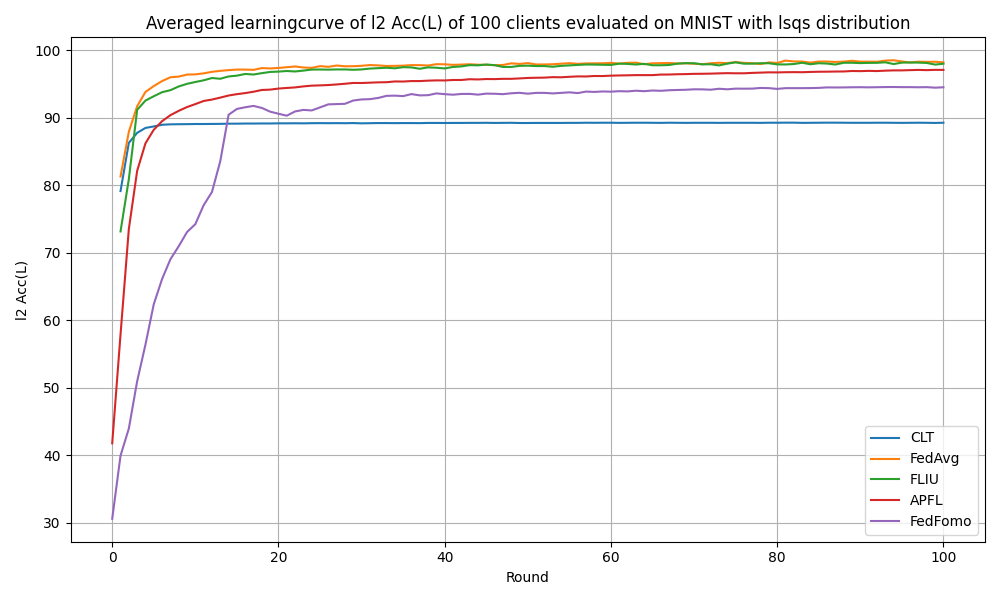}}
    \caption{Comparison of the trajectories of $\mathbf{Acc}\left(L\right)$ at stage $L_2$ of CLT, FedAvg, APFL, FedFomo and FLIU on MNIST with LSQS for T=100 rounds.}\label{fig:eval2_MNIST_lsqs_learningcurve}
\end{figure}
\begin{figure}[h]
    \centering
    {\includegraphics[width=\linewidth]{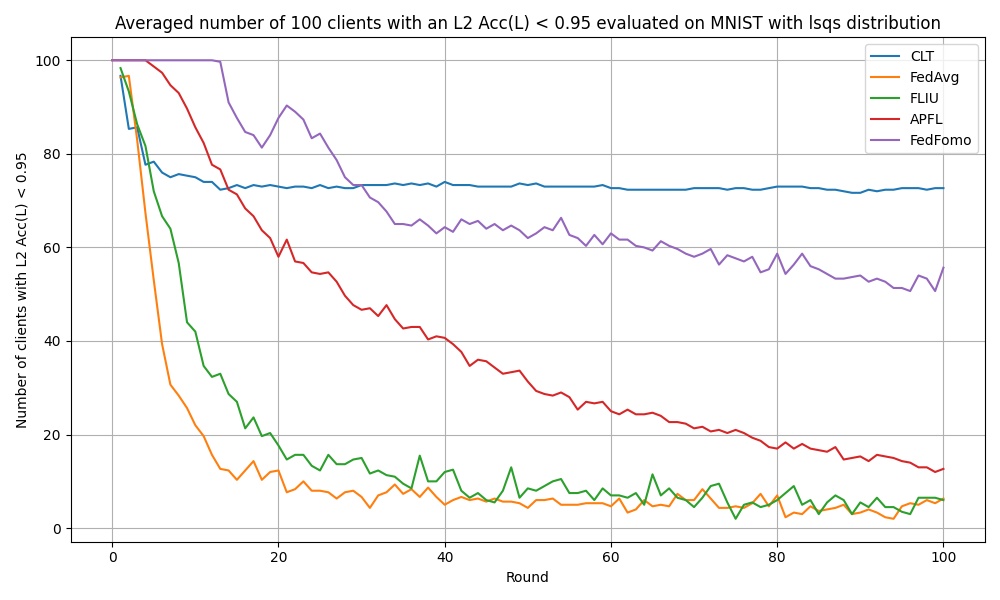}}
    \caption{Comparison of the trajectories of the number of clients with $\mathbf{Acc}\left(L\right)$ lower than 95\% at stage $L_2$ of CLT, FedAvg, APFL, FedFomo and FLIU on MNIST with LSQS for T=100 rounds.}\label{fig:eval2_MNIST_lsqs_rho}
\end{figure}
As illustrated in Figure \ref{fig:eval2_MNIST_lsqs_learningcurve}, the learning curve provides further insight by demonstrating the exemplary development of $L_2$ $\mathbf{Acc_k}\left(L\right)$.
As demonstrated, all algorithms rapidly attain a certain level of performance.
However, it turned out that the average number of rounds required for FedAvg and FLIU to achieve 90\% accuracy is three, whereas APFL and FedFomo require seven and fourteen rounds, respectively (cf. Fig. \ref{fig:eval2_MNIST_lsqs_learningcurve}).
This finding suggests that a considerable number of clients derive less immediate benefit from the federated system.

In order to conduct a more thorough investigation, an additional evaluation was performed to ascertain the number of clients on each round whose $L_2$ $\mathbf{Acc_k}\left(L\right)$ is less than a certain performance level $\rho$.
This evaluation was conducted to determine the number of clients who benefit from the approach and the number of clients for whom it is less effective.
As illustrated in Figure \ref{fig:eval2_MNIST_lsqs_rho}, FLIU, predominantly FedAvg, enhances the majority of its clients to a $L_2$ $\mathbf{Acc_k}\left(L\right)$ greater than 95\%.
Consequently, only a limited number of clients have a lower accuracy level (still greater than 85\%).
In comparison to that, APFL and FedFomo demonstrate deficiencies in their ability to maintain consistent performance levels, since many clients attain an $L_2$ $\mathbf{Acc_k}\left(L\right)$ of at least 95\% at a subsequent point or not at all.
In this particular context, it is noteworthy that APFL, which attains an $L_2$ $\mathbf{Acc}\left(L\right)$ comparable to that of FedAvg and FLIU after 100 rounds, possesses a significantly larger client base, who do not reap the same benefits as the average client.
However, in comparison with CLT, all FL approaches achieve superior accuracies and have the capacity to elevate a greater number of clients to higher performance levels. Specifically, FedAvg and FLIU demonstrate the capability to enhance a substantial number of individuals while simultaneously ensuring that only a limited number are left behind.
It is also noteworthy that the vast majority of approaches demonstrate a high degree of proficiency in handling the QS environment, despite this subject is not extensively addressed in their respective papers.

\section{Conclusion}\label{sec:Conclusion}

The empirical results obtained clearly illustrate the conflict between local performance and generalization in a first step.
The aggregation of local models has been demonstrated to promote better generalization, thereby enabling individual clients to possess a more robust model.
However, this aggregation concurrently reduces their performance on locally trained conditions.
Consequently, the personalization of individual models through local adaptation appears to be indispensable.
Nonetheless, this will in turn result in a decline in acquired generalization again.
Our proposed FLIU technique has obtained performance across a range of environmental settings in terms of local performance and out-of-distribution generalization has been demonstrated to be competitive with FedAvg and some PFL approaches, such as APFL and FedFomo.
In certain instances, it has even surpassed them.
Even though more experiments additionally with more PFL approaches could be conducted.
In general, and particularly for data environments that are relatively close to IID or uniformly distributed, our approach favors a smaller $\gamma$.
In cases where the distribution is particularly extreme and exhibits imbalance and non-IID characteristics, the importance of a larger $\gamma$ becomes more pronounced.
This outcome demonstrates the efficacy of our adaptive function, as this method appears to strike a good balance between local performance and generalization, regardless of the data environment.
It is noteworthy that APFL frequently achieved superior outcomes in experiments with a mere 10 clients, but could not continue its superiority when considering 100 clients.
Additionally, FedAvg has proven capable of handling data environments with skewness in label and/or quantity, such as LS, QS or LSQS, even though it struggles with extreme skewness, such as PATH.

Furthermore, despite the general challenges of FL, there are still many areas to investigate regarding the performance and robustness of FL and PFL in complex yet more realistic data environments.
Our work examined only one aspect of robustness, namely out-of-distribution generalization.
However, other aspects exist that are not stressed here but have an important impact on the resilience of ML methods, such as noise and bias in data or client manipulation.
Nevertheless, the present experiments have clearly demonstrated that there are still insufficiently researched effects of the various FL approaches on the trade-off between local performance and out-of-distribution generalization.
Regarding FLIU, it seems that our $\gamma$ formula (\ref{eq:gamma}) could use some modification.
For example, we could modify the range of $\gamma$ from $0.1$ to $0.5$.
Our experiments show that a higher personalization factor comes too close to CLT, which performed poorly in all cases.
However, further analysis of $\gamma \in \left( 0.5, 0.75 \right)$ would provide more evidence of this.
We pursue the extension of our research work by the addition of alternative adaptation methods for the personalization factor $\gamma$.
In further work, we aim to expand our approach into a hierarchical construct in order to gain further opportunities for adaptation and increased robustness and resilience, e.g., through intelligent clustering methods.

\bibliographystyle{IEEEtran}
\bibliography{IEEEabrv,references}

\end{document}